\documentclass[10pt,twocolumn,letterpaper]{article}

\usepackage{cvpr}              

\usepackage{url}
\usepackage{amsthm,amsmath,amssymb}
\usepackage{mathrsfs}
\usepackage{multirow}

\definecolor{cvprblue}{rgb}{0.21,0.49,0.74}
\usepackage{graphicx}
\usepackage{enumitem}
\usepackage[pagebackref,breaklinks,colorlinks,allcolors=cvprblue]{hyperref}
\usepackage[accsupp]{axessibility}

\def\paperID{2187} 
\def\confName{CVPR}
\def\confYear{2025}

\title{PatchVSR: Breaking Video Diffusion Resolution Limits with \\Patch-wise Video Super-Resolution}

\author{Shian Du\textsuperscript{$\rm 1^{\ast\ddagger}$}, Menghan Xia\textsuperscript{$\rm 2^{\dagger\ddagger}$}, Chang Liu\textsuperscript{\rm 1}, Xintao Wang\textsuperscript{\rm 2}, Jing Wang\textsuperscript{\rm 3}, Pengfei Wan\textsuperscript{\rm 2}, \\Di Zhang\textsuperscript{\rm 2}, Xiangyang Ji\textsuperscript{$\rm 1^{\dagger}$} \\
\textsuperscript{\rm 1}Tsinghua University, \textsuperscript{\rm 2}Kling Team, Kuaishou Technology, \\ \textsuperscript{\rm 3}Beijing Institute of Technology \\
}

\newcommand\nnfootnote[1]{%
  \begin{NoHyper}
  \renewcommand\thefootnote{}\footnote{#1}%
  \addtocounter{footnote}{-1}%
  \end{NoHyper}
}

\begin{document}

\maketitle
\nnfootnote{$\ast$ This work was conducted during the author's internship at Kling Team, Kuaishou Technology.}
\nnfootnote{$\ddagger$ Equal contribution. $\dagger$ Corresponding author.}

\begin{abstract}
Pre-trained video generation models hold great potential for generative video super-resolution (VSR). 
However, adapting them for full-size VSR, as most existing methods do, suffers from unnecessary intensive full-attention computation and fixed output resolution. To overcome these limitations, we make the first exploration into utilizing video diffusion priors for patch-wise VSR.
This is non-trivial because pre-trained video diffusion models are not native for patch-level detail generation. To mitigate this challenge, we propose an innovative approach, called \textit{PatchVSR}, which integrates a dual-stream adapter for conditional guidance. The patch branch extracts features from input patches to maintain content fidelity while the global branch extracts context features from the resized full video to bridge the generation gap caused by
incomplete semantics of patches. Particularly, we also inject the patch's location information into the model to better contextualize patch synthesis within the global video frame. Experiments demonstrate that our method can synthesize high-fidelity, high-resolution details at the patch level. 
A tailor-made multi-patch joint modulation is proposed to ensure visual consistency across individually enhanced patches. 
Due to the flexibility of our patch-based paradigm, we can achieve highly competitive 4K VSR based on a 512$\times$512 resolution base model, with extremely high efficiency.
\end{abstract}    
\section{Introduction}
\label{sec:introduction}

Video super-resolution (VSR) has long been a challenging research problem in the field of computer vision. With the advent of deep learning, significant advancements have been made in end-to-end VSR models trained on curated datasets. However, due to limited model capacity and data coverage, these models typically achieve sharp object structures or coarse textures but struggle to synthesize faithful details.
Recently, the emergence of diffusion generative models has enabled open-domain video generation of remarkable quality. The generative capabilities of these foundation models present new opportunities for video super-resolution techniques~\cite{zhou2024upscale,he2024venhancer}. Nevertheless, existing base models are generally constrained to generating videos of fixed sizes due to their inherent architecture. This limitation is inherited in their downstream super-resolution models, making it challenging for them to support arbitrary resolution outputs.

This study is motivated by the insight that feature attention in super-resolution models is more locally focused compared to generation models. This localization is reasonable, as super-resolution tasks have a low-resolution video as a reference, allowing details to be generated based on neighborhood semantics rather than requiring intensive global coherence considerations.
To leverage this insight, we propose a patch-based strategy for video super-resolution. The concept is straightforward and intuitive: we split the input video into patches sized to be compatible with the pre-trained base model, enhance each patch individually, and then combine them to produce the final result. This approach enables the achievement of super-resolution results at arbitrary resolutions, independent of the base model's inherent resolution constraints.

The key challenge lies in that video generation-based models are trained on complete video frames and face noticeable performance degradation for patch-level generation.
To adapt a pre-trained video generation model for generating patch-level details in super-resolution, we propose an innovative approach, called \textit{PatchVSR}, which integrates a dual-stream adapter for conditional guidance. The local branch extracts features from the input patches to guide the base model for detail synthesis adhering to the input, while the global branch captures contextual information from the entire input frame to bridge the generation gap caused by incomplete semantics of patches.
To make the context guidance more customized, the global branch also incorporates a binary mask indicating the target patch position as input. Additionally, we employ LoRA technology to fine-tune the base model, further improving its adaptability for generating patch-level details. 

Experimental results demonstrate that our method effectively adapts the generative ability of the base model to produce faithful details at the patch level. Specifically, to alleviate the boundary inconsistency across patches, we propose a training-free multi-patch joint modulation scheme, where auxiliary bridging patches are created to allow feature interaction across patches during generation.
Thanks to the flexibility of PatchVSR, our method, based on a pre-trained $512\times 512$ resolution video base model, supports super-resolution video generation up to 4K resolution, while significantly outperforming current state-of-the-art methods in terms of fidelity and detail realism. Notably, compared to existing methods that operate on full-frame videos, our approach processes relatively smaller patches, without full-range attention computation, demonstrating a significant efficiency advantage.

Our contributions are summarized as follows:
\begin{itemize}
\itemindent=5pt
    \item We introduce an innovative approach to overcome the resolution limitations of using pre-trained video diffusion models for VSR. Our method achieves any-resolution VSR output based on a fixed-resolution base model.
    \item We propose a novel dual-branch adapter paradigm that effectively harnesses the capabilities of pre-trained video models for patch-level detail generation.
    \item Through extensive experimentation, we demonstrate the significant superiority of our method in terms of faithful detail generation and computational efficiency.
\end{itemize}

\section{Related Work}
\label{sec:related-work}

\subsection{Text-to-video Generation}
With the widespread use of pre-trained text-to-image models \cite{rombach2022high,zhang2023adding,mou2024t2i,hertz2022prompt,du2024efficient,gal2022image} and the preparation of large-scale text-video pairing datasets \cite{chen2024panda,nan2024openvid}, substantial efforts have been made to train a large-scale text-to-video model. Instead of training from scratch, some methods \cite{wang2023modelscope,blattmann2023stable,zhang2024show,wu2023tune} focus on inserting temporal modules and fine-tuning them or all parameters on video data. To avoid catastrophic forgetting and degradation on visual quality, they often adopt image-video joint training strategy \cite{chen2023videocrafter1,chen2024videocrafter2,xing2025dynamicrafter}. Recently, with Diffusion Transformer (DiT) \cite{peebles2023scalable} being proposed, it has replaced the traditional 3D UNet architecture used by previous approaches with better generation quality and scalability, and has become the dominant framework for text-to-video generation \cite{yang2024cogvideox,ma2024latte,liu2024sora}. Although achieving high visual quality, extending them to higher-resolution is non-trivial since it requires substantial training resource and high-quality high-resolution video dataset, which is unavailable by this moment. To bypass this difficulty, current methods often adopt a cascade framework \cite{wang2023lavie,zhang2023i2vgen,guo2024make}: the low-resolution video is first generated by the base model using text prompt, and the refinement model outputs high-resolution video through a noising-denoising process \cite{meng2021sdedit} by taking the low-resolution video as condition. However, existing methods are limited to generate full video at once, which still demands a high training overhead. At the same time, the low inference efficiency and high memory usage also hinder its downstream applications.

\begin{figure*}[!t]
    \centering
    \includegraphics[width=1\linewidth]{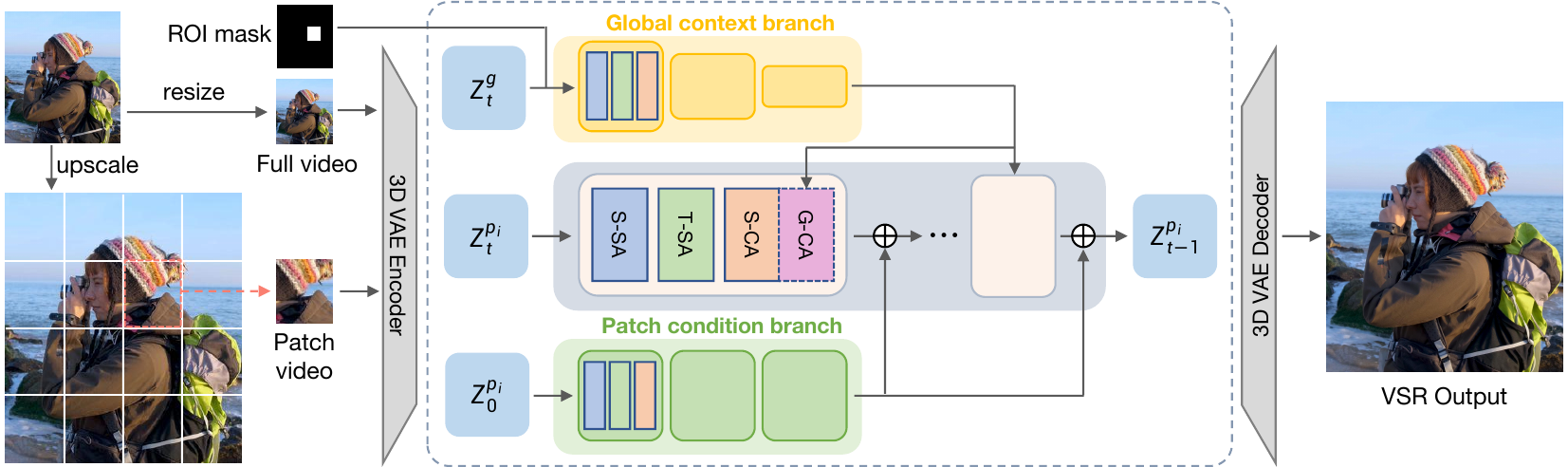}
    \vspace{-1.0em}
    \caption{Flowchart of our \emph{PatchVSR}. Building upon a pre-trained latent T2V model, we incorporate a patch condition branch and a global context branch. These branches extract features from partitioned video patches and the resized full video together with a binary mask that indicates the location of the ROI patch, respectively. Particularly, local patch features are added to the output of each block, while the global context feature is fused with the backbone feature through newly introduced cross-attention modules (\textit{G-CA}). For simplicity, we have omitted other conditional inputs such as text prompts and time steps from this diagram. The processed patches are fused via a joint modulation scheme to produce a coherent super-resolution video.
    }
    \label{fig:overview}
    \vspace{-1.0em}
\end{figure*}

\subsection{Video Super-Resolution}
Video super-resolution aims at restoring high-resolution (HR) video frames from its degraded low-resolution (LR) counterparts and enhancing its visual quality. Traditional methods often assume a fixed degradation process and their performance deteriorates significantly when handling AI-generated and real-world videos. Due to the lack of real-world LR-HR video data pair for training, RealVSR \cite{yang2021real} collects LR-HR video data pair by utilizing the multi-camera system of iPhone. Nevertheless, the sequences are captured by a single device, which only contains camera-specific degradations. To avoid burdensome data collection, other methods \cite{chan2022investigating, zhou2024upscale,jeelani2023expanding} use stochastic combinations of multi-order degradation processes to simulate real-world degradation processes. While performing more robust when handling real-world cases, due to the absence of deep generative prior, it is challenging for traditional CNN-based \cite{chan2021basicvsr,chan2022basicvsr++,chan2022investigating,wang2019edvr,isobe2020video} and transformer-based \cite{cao2021video,liu2022learning,shi2022rethinking,geng2022rstt} methods to generate photo-realistic details and textures. By utilizing a pre-trained image diffusion model like Stable Diffusion (SD) $\times 4$ Upscaler, Lavie-SR \cite{wang2023lavie} and Upscale-A-Video \cite{zhou2024upscale} circumvent the need for exhaustive training from scratch and exhibit improved performance in fixed spatial scale within a short sequence. To improve temporal consistency for long sequence, I2VGen-XL \cite{zhang2023i2vgen} and VEnhancer \cite{he2024venhancer} utilize a pre-trained low-resolution video generative model as the base model, and directly fine-tune the model parameters using high-resolution video dataset following noising-denoising manner \cite{meng2021sdedit}. However, the aforementioned methods are all limited to generate full video at once, which requires expanding the training resolution of pre-trained model, bringing low training efficiency and slow convergence speed. Meanwhile, high resolution pre-trained video generative models are currently unavailable to access, hindering the scalability of video super-resolution methods with deep generative prior. To tackle this problem, we make the first exploration of utilizing pre-trained video generative model for patch-wise video super-resolution. By maintaining the training resolution of the pre-trained model, our approach not only avoids significant training overhead, making it possible to generate 4K resolution videos using a $512\times512$ resolution pre-trained model. At the same time, our method significantly improves inference efficiency and reduces memory usage, facilitating downstream applications of video super-resolution methods.

\section{Method}
\label{sec:method}



We aim to develop a video super-resolution (VSR) framework designed to restore degraded videos of arbitrary resolutions by generating visually coherent and structurally faithful details. Our primary innovation involves strategically adapting pre-trained text-to-video (T2V) diffusion models—powerful generative tools typically constrained to fixed output size (denoted as $s=h \times w$)—through a curated conditioning paradigm. To bypass their inherent token-length limitations, we introduce a patch-based processing strategy: the input video is decomposed into $s$-sized patches, each independently enhanced using the diffusion model, and then reintegrated into a unified high-resolution output through optimized spatial-temporal fusion.
This approach is grounded in the distinct characteristics of VSR compared to video generation. While video synthesis demands global coherence across full-frame contexts, super-resolution prioritizes localized detail restoration atop existing structural content.

Specifically, we augment the base model with two conditional branches: a patch branch that extracts features from input patches, and a global branch that extracts context from the full video. This global context mainly aids in generating patch-level details from the T2V model that was originally trained on full-frame videos. An overview of the model architecture is illustrated in Fig.~\ref{fig:overview}.
Given a low-resolution image $\mathbf{V}_{l} \in \mathbb{R}^{F\times H\times W\times 3}$ and the target upscale factor $k$, our process begins by upscaling $\mathbf{V}_{l}$ to the desired resolution $\mathbf{V}_{l}^{\uparrow} \in \mathbb{R}^{F\times \lfloor{H\cdot k}\rfloor\times \lfloor{W\cdot k}\rfloor\times 3}$ using bicubic interpolation per frame. We then partition $\mathbf{V}_{l}^{\uparrow}$ into patches $\{\mathbf{P}_i \in \mathbb{R}^{F \times h \times w \times 3}\}_{i=1}^N$, where the patch dimensions $(h,w)$ match the generation size of the pre-trained T2V model. Additionally, we resize the full video to $\mathbf{G} \in \mathbb{R}^{F \times h \times w \times 3}$ to serve as global guidance. Finally, both $\mathbf{P}_i$ and $\mathbf{G}$ are input into the VSR model for patch-level detail enhancement, and multi-patch joint modulation is involved to ensure coherence across patches.
This formulation ensures compatibility with the pre-trained T2V model while allowing for flexible processing of videos of arbitrary sizes.

\subsection{Preliminary: Video Diffusion Base Model}
\label{subsec:preliminary}

Our model is implemented on top of a pre-trained T2V diffusion model, which adopts the Rectified Flow framework~\cite{scaling_esser_2024} for the noise schedule and denoising process. The forward process is defined as straight paths between data distribution and a standard normal distribution, i.e.
\begin{equation}\label{eq:forward}
    z_t = (1-t)z_0 + t\epsilon,
\end{equation}
where $\epsilon \in \mathcal{N}(0,I)$ and $t$ denotes the iterative timestep.
To solve the denoising processing, we define a mapping between samples $z_1$ from a noise distribution $p_1$ to samples
$z_0$ from a data distribution $p_0$ in terms of an ordinary differential equation (ODE), namely:
\begin{equation}\label{eq:ODE}
dz_t=v_{\Theta}(z_t,t)dt, 
\end{equation}
where the velocity $v$ is parameterized by the weights $\Theta$ of a neural network. For inference, we employ Euler discretization for Eq.~\ref{eq:ODE} and perform discretization over the timestep interval at $[0, 1]$, starting at $t=1$. We then processed with iterative sampling with: $z_{t_i}=z_{t_j} + v_{\Theta}(z_{t_j},t_j) * (t_j-t_i)$.

For consideration of computing efficiency, the mapping is formulated in a latent space constructed with a 3D Variational Auto-Encoder (VAE)~\cite{kingma2014autoencoding}. The core is a Transformer-based diffusion model (DiT)~\cite{dit}, where each block is instantiated as a sequence of spatial attention, temporal attention, and cross-attention modules. Text prompts and other micro parameters (like time step, aspect ratio, etc) are used as generative conditions. Due to the full-attention property of Transformer, our model only support the generation of a fixed number of tokens as it was trained with, i.e. $512\times512$ of diverse aspect-ratio.
This is a common characteristic of almost all existing diffusion models for image or video generation~\cite{podell2023sdxl,chen2023videocrafter1,chen2024videocrafter2,yang2024cogvideox}, and extending pre-trained diffusion models for higher resolution generation remains an open research topic~\cite{he2024scalecrafter,guo2024make}. For VSR as a downstream task, we propose a patch-wise processing strategy to break this resolution limit.

\subsection{Patch Video Diffusion with Global Guidance}
\label{subsec:conditonal_gen}

To adapt a pre-trained T2V model for patch-level VSR, we must consider two key aspects: (i) enabling the input video to guide the base model's denoising process and (ii) bridging the gap between the base model's capabilities in generating full-frame content and patch-level content synthesis requirements. We achieve this through a dual-branch adapter, as detailed below.

\paragraph{Patch Condition Branch.}
There are generally two approaches to incorporating visual conditions into pre-trained generative models: concatenating with noisy data in the noise space~\cite{zhou2024upscale,rombach2022high}, or fusing with the backbone feature through an adapter branch~\cite{wang2024exploiting,he2024venhancer}. The former involves modifying the first layers' channel count and necessitates base model fine-tuning, which is typically more data and computationally intensive. In contrast, the latter is more lightweight, as only a newly added branch requires training for feature adaptation. We adopt a hybrid solution that combines elements of both approaches. First, we employ an adapter comprising several Transformer blocks to extract features from the input video patch $\mathbf{P}_i$ and inject them into the base model's blocks. Additionally, considering that the base model was trained for full video generation, we need to adapt its embedded knowledge for patch-level generation. To this end, we allow the base model to be fine-tuned using a Low-Rank Adaptation (LoRA)~\cite{hu2022lora}. An ablation study in Sec.~\ref{subsec:ablation} demonstrates the advantages of this design.

\paragraph{Global Context Branch.}
To further guide content generation in a natural manner, we incorporate a global branch that extracts context from the full input video. This branch utilizes a Transformer-based encoder architecture that reduces the number of tokens to one quarter. The resulting context features are fused with each block of the base model through cross-attention $\{\mathbf{Q},\mathbf{K}_g,\mathbf{V}_g\}$, where the $\mathbf{Q}$ is shared with the existing text-prompt cross-attention $\{\mathbf{Q},\mathbf{K}_t,\mathbf{V}_t\}$. Specifically, the input is constructed by concatenating the resized full video $\mathbf{V}_g$ (to accommodate a fixed number of tokens) with a binary map $\mathbf{M}_i$ indicating the target patch's $\mathbf{P}_i$ location within it. Our experiments show that the global context branch not only enhances the vividness of generated patch-level details but also improves consistency across patches.

\begin{figure}
    \centering
    \includegraphics[width=\linewidth]{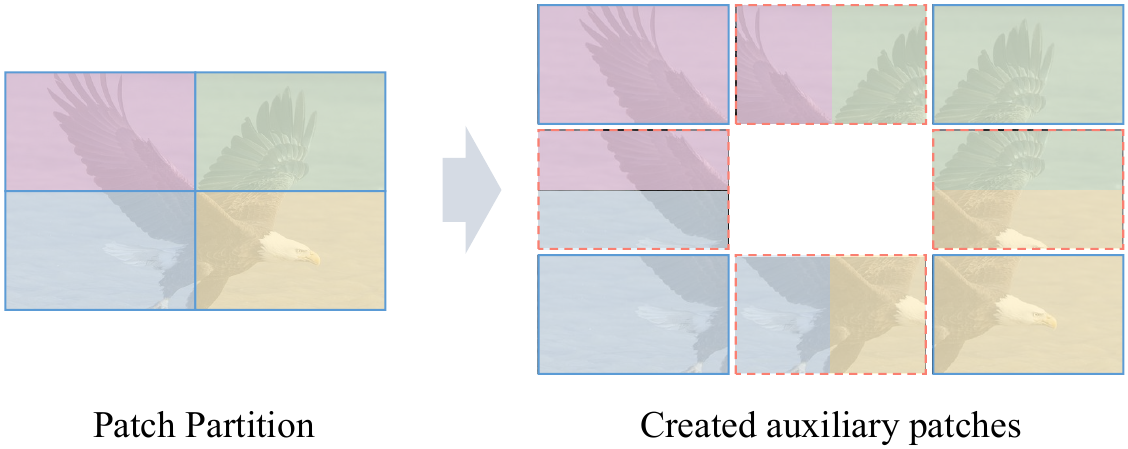}\vspace{-0.5em}
    \caption{Patch Partition Visualization. The input video is divided into non-overlapping segments, as the solid blue boxes mark. For joint modulation, auxiliary patches are created, indicated by the red dashed boxes, resulting in an overlapping ratio of $50\%$.}
    \label{fig:patch_partition}
\label{fig:stitched_video}\vspace{-1.5em}
\end{figure}

\subsection{Multi-Patch Joint Modulation}
\label{subsec:multi-patch}

\begin{figure*}[!t]
    \centering
    \includegraphics[width=\linewidth]{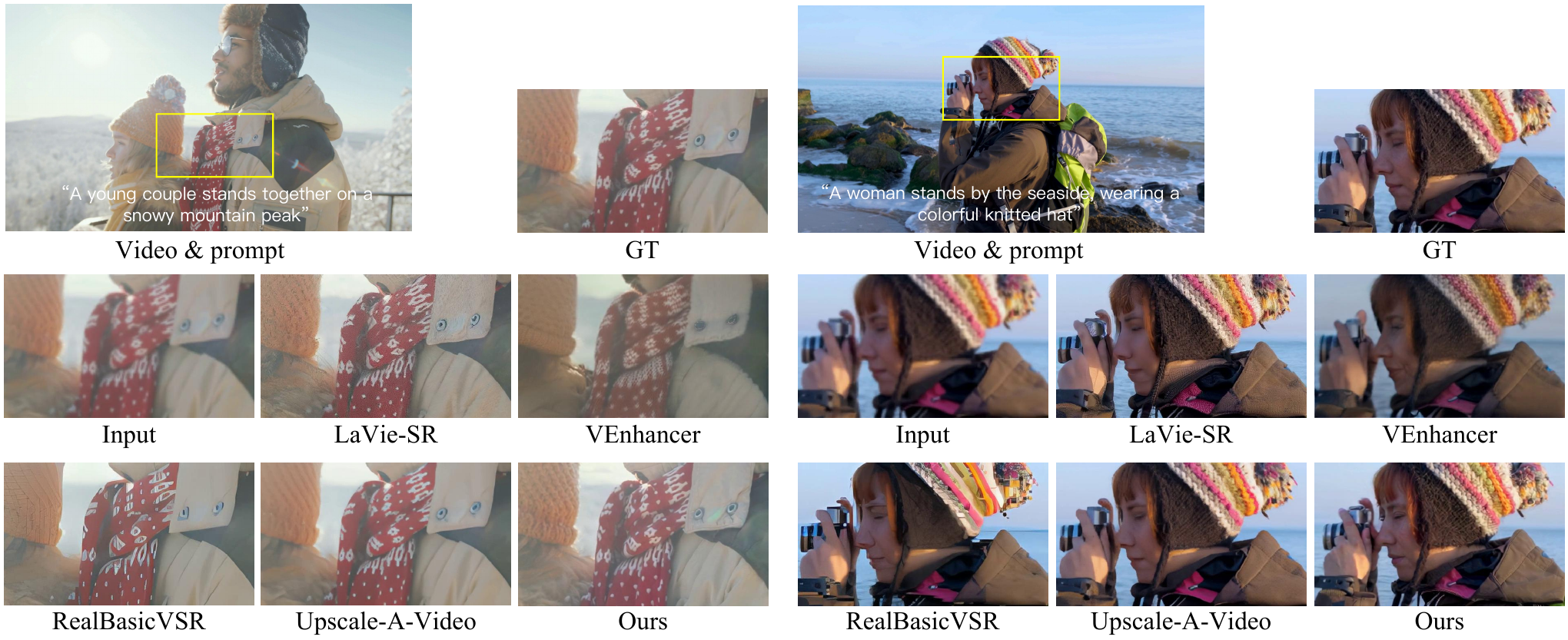}\vspace{-0.5em}
    \caption{Qualitative comparisons on SynVideo30 (test). These videos $\times 4$ super-resolution results. \textbf{Zoom in for best view.}}
    \label{fig:qualitative-synthetic}
    \vspace{-0.5em}
\end{figure*}

While the global guidance shared by patches is critical for maintaining color consistency, tonal uniformity, and semantic coherence of details, naive stitching of independently enhanced patches often introduces visible seams. These boundary artifacts primarily stem from the stochastic nature of the generative process when hallucinating textures on degraded or underdetermined regions. This is a fundamental challenge in ill-posed inverse problems where multiple high-resolution solutions may satisfy the constraints of a single low-resolution input.

To mitigate this problem, we introduce inter-patch modulation during the detail generation process. Specifically, a training-free inference strategy called multi-diffusion~\cite{bar2023multidiffusion} is adapted to solve our problem. This approach involves splitting the original video into overlapping patches and updating the overlapping regions by averaging contributions from all relevant patches at each denoising step.
As illustrated in Fig.~\ref{fig:patch_partition}, instead of directly splitting the video into overlapping patches, we first partition it into non-overlapping segments, and then construct auxiliary patches by combining the halves of adjacent patches. This creates a new set of patch groups with an overlapping ratio of $50\%$.
Our experiment shows that directly taking the average in overlap regions suffers from artifacts of black holes or seamlines, as evidenced in Fig.~\ref{fig:patch_fusion} (middle). 
Instead, when fusing base patches and its related auxiliary patches, we manage to gradually reduce its influence of auxiliary patches from its center splitting line to two sides by assigning spatial weight maps to them.
This method effectively reduces visible seams between patches and promotes overall consistency. A comparative result demonstrating the effectiveness of this approach is presented in Fig.~\ref{fig:patch_fusion}.

\section{Experiments}
\label{sec:experiments}

\begin{figure*}[!t]
    \centering
    \includegraphics[width=\linewidth]{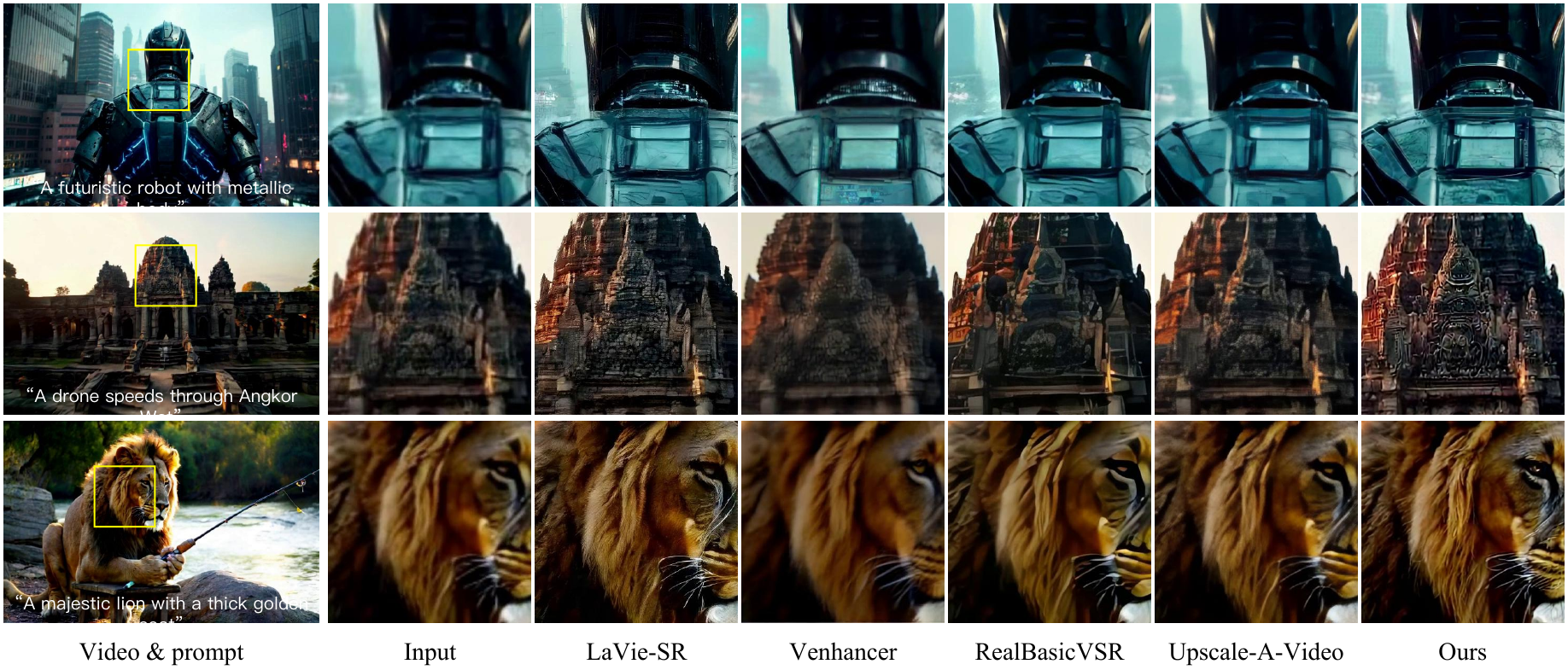}\vspace{-0.5em}
    \caption{Qualitative comparisons on VideoGen30 (test). These videos are $\times 4$ super-resolution results. \textbf{Zoom in for best view.}}
    \label{fig:qualitative-aigc}
\end{figure*}

\begin{table*}[!t]	
	\centering
	\begin{tabular}{l|l|cccc|c}
		\toprule
		Datasets & Metrics & RealBasicVSR \cite{chan2022investigating} & LaVie-SR \cite{wang2023lavie} & Upscale-A-Video \cite{zhou2024upscale} & VEnhancer \cite{he2024venhancer} & Ours \\
		\midrule
		REDS30 & PSNR $\uparrow$ & 29.698 & 28.735 & \textcolor{red}{31.257} & 28.544 & \textcolor{blue}{30.111} \\
        & SSIM $\uparrow$ & \textcolor{blue}{0.656} & 0.622 & \textcolor{red}{0.661} & 0.611 & 0.647 \\
        & LPIPS $\downarrow$ & 0.334 & 0.407 & \textcolor{blue}{0.328} & 0.421 & \textcolor{red}{0.326} \\
        & DOVER $\uparrow$ & 0.305 & 0.318 & 0.310 & \textcolor{red}{0.344} & \textcolor{blue}{0.323} \\
        & MUSIQ $\uparrow$ & 40.587 & 43.577 & 42.575 & \textcolor{blue}{44.533} & \textcolor{red}{44.956} \\
		& Aesthetics $\uparrow$ & 0.422 & \textcolor{blue}{0.486} & 0.471 & 0.482 & \textcolor{red}{0.493} \\
		\hline
		SynVideo30 & PSNR $\uparrow$ & \textcolor{red}{33.507} & 31.433 & \textcolor{blue}{33.432} & 28.856 & 30.857 \\
        & SSIM $\uparrow$ & \textcolor{red}{0.776} & 0.716 & 0.728 & 0.697 & \textcolor{blue}{0.732} \\
        & LPIPS $\downarrow$ & \textcolor{blue}{0.185} & 0.194 & 0.205 & 0.199 & \textcolor{red}{0.183} \\
        & DOVER $\uparrow$ & 0.584 & \textcolor{red}{0.612} & 0.529 & 0.550 & \textcolor{blue}{0.597} \\
        & MUSIQ $\uparrow$ & 49.557 & 48.146 & \textcolor{blue}{49.839} & 43.538 & \textcolor{red}{50.695} \\
		& Aesthetics $\uparrow$ & 0.496 & \textcolor{blue}{0.509} & 0.494 & 0.503 & \textcolor{red}{0.520} \\
        \hline
		VideoGen30 & DOVER $\uparrow$ & 0.588 & \textcolor{red}{0.609} & 0.510 & 0.512 & \textcolor{blue}{0.590} \\
        & MUSIQ $\uparrow$ & 47.564 & \textcolor{blue}{48.329} & 46.899 & 46.930 & \textcolor{red}{50.559} \\
		& Aesthetics $\uparrow$ & 0.547 & \textcolor{blue}{0.600} & 0.574 & 0.551 & \textcolor{red}{0.602} \\
		\bottomrule
	\end{tabular}\vspace{-0.5em}
     \caption{Quantitative comparisons on different VSR benchmarks from synthetic (REDS30, SynVideo30) and AIGC (VideoGen30) data. \textcolor{red}{Red} and \textcolor{blue}{blue} numbers indicate the \textcolor{red}{best} and \textcolor{blue}{second best} results respectively.}
	\label{tab:quantitative-comparisons}\vspace{-1.0em}
\end{table*}

To verify the superiority of PatchVSR, we compare it with SOTA methods, including VEnhancer \cite{he2024venhancer}, Upscale-A-Video \cite{zhou2024upscale}, LaVie-SR \cite{wang2023lavie} and RealBasicVSR \cite{chan2022investigating}. We also make ablation study to analyze the effectiveness of our proposed designs. Besides, some extended analysis is conducted to demonstrate the capability of our method.

\subsection{Implementation Details}

\paragraph{Training Datasets} We train our model using 460K self-collected high-quality video-text pairs, with each resolution from $1024\times1024$ to 2K. The text prompts are all captioned using LLAVA captioner \cite{liu2024visual}, and encoded by T5 text encoder \cite{raffel2020exploring} with no more than 256 tokens.
\vspace{-0.5em}
\paragraph{Testing Datasets.} We evaluate the VSR methods on two kinds of dataset, i.e. downscale synthetic videos and AIGC videos. For synthetic videos, since our employed base model is fixed to generate 77-frames video, we utilize a public dataset: REDS30 \cite{nah2019ntire} for fair comparisons, which contains 100 frames. To evaluate higher resolution, we self-collect 30 high-quality 2K resolution videos, each containing more than 150 frames. Additionally, we collect and test on an AIGC dataset, which comprises 30 AI-generated videos from public text-to-video generative models \cite{zeng2024dawn}.
\vspace{-0.5em}
\paragraph{Training Details.} Our model is trained on 16 NVIDIA H800-80G GPUs with a total batch size of 64. AdamW \cite{loshchilov2017decoupled} is used as the optimizer with a learning rate of $10^{-5}$. The text prompt is randomly replaced by a null prompt with $10\%$ probability. To obtain the LR-HR training data pair, the HR video is first down-sampled using bilinear interpolation with down scale $s$ and up-sampled back to the original resolution. To match the size of the pre-trained model, we randomly crop a $512\times512$ area from the up-sampled LR video with aspect ratios from 0.5 to 2.0. To enhance the robustness of our model to different degradation scenarios, we utilize the noise augmentation technique by injecting noise into the input latent using a diffuse process. The noise timestep is randomly selected from 200 to 300, aiming at preserving the structure of the input video. We have additionally encoded noise timestep, downscale factor, and crop location as conditions for the model. We have replaced the text prompt with a fixed prompt for the patch condition branch, with the base model and global context branch using text prompts. To provide initialization weight, we followed the structure of the pre-trained model and sampled the layer index from it at equal intervals.
\vspace{-0.5em}
\paragraph{Inference Details.} We perform 50 PNDM \cite{liu2022pseudo} sampling steps with classifier-free guidance scale $5.0$. We have also used timestep shift \cite{esser2024scaling} with shift value $5.0$. 
\vspace{-0.5em}
\paragraph{Evaluation Metrics.} To evaluate the fidelity of our method on synthesized videos, we utilize PSNR, SSIM, and LPIPS \cite{zhang2018unreasonable}. For visual quality, we conduct our evaluation of commonly used quality assessment metrics MUSIQ \cite{ke2021musiq} and DOVER \cite{wu2023exploring}. Furthermore, we also include \textit{aesthetic score} from a video generation benchmark VBench \cite{huang2024vbench} for more comprehensive comparisons.

\subsection{Comparison with SOTA methods}
\label{subsec:comparison}

\begin{figure}[t]
    \centering
    \includegraphics[width=\linewidth]{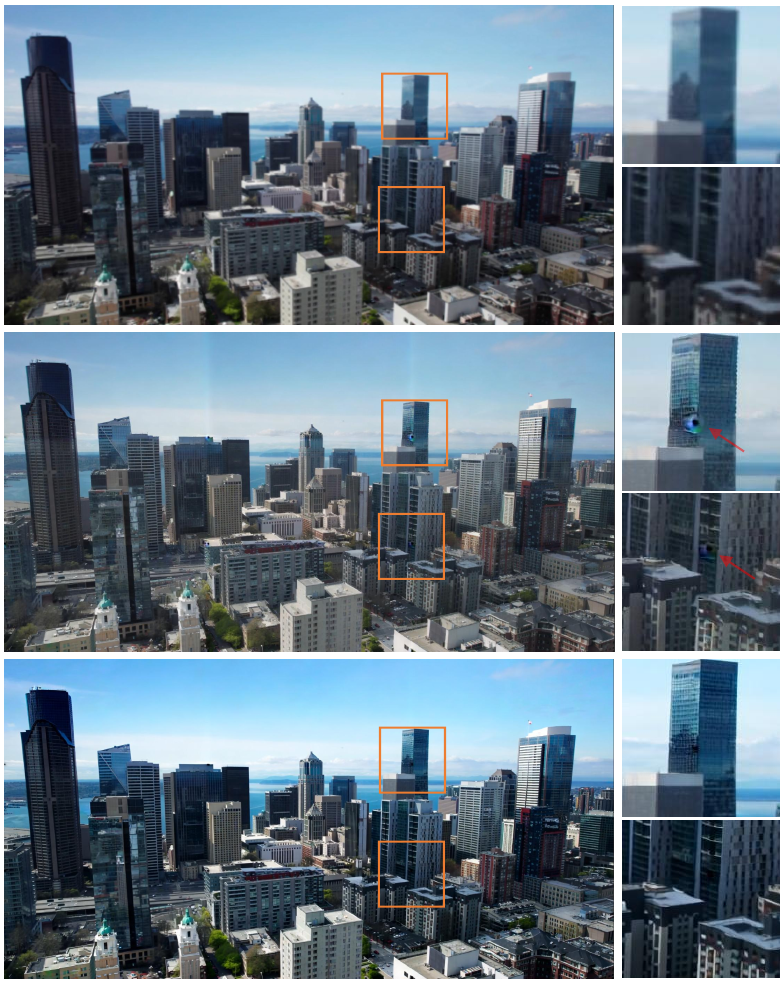}\vspace{-0.5em}
    \caption{Visual evaluation on multi-patch joint modulation. Here the super-resolved videos of 2K resolution are used for comparison, where $3\times 3=9$ patches are involved. The top, middle, bottom denote input video, result of stitching latent of patches, and result of ours respectively.}
    \label{fig:patch_fusion}\vspace{-1.0em}
\end{figure}

\paragraph{Qualitative Results.}
The visual results of synthetic and AIGC videos are presented in Fig.~\ref{fig:qualitative-synthetic} and Fig.~\ref{fig:qualitative-aigc} respectively. It is observed that our method significantly outperforms existing CNN-based and diffusion-based models in both details generation and temporal consistency (Readers are recommended to check our provided video results). Specifically, our method not only successfully recovers the details of the red silk scarf in Fig.~\ref{fig:qualitative-synthetic}, but also adds more details to the ``Angkor Wat'' as shown in Fig.~\ref{fig:qualitative-aigc}, which others tend to generate blurry or unnatural results. 
In Fig.~\ref{fig:patch_fusion}, we also illustrate full VSR results achieved through the multi-patch joint modulation scheme. It can be seen that creating auxiliary patches greatly alleviates the seam problem and significantly reduces the color difference between different patches, while further improving the contrast of the results.
To show the advantage of our multi-patch joint modulation, we compare with a baseline method: stitching the patches in latent space and decoding the big latent together, which suffers obvious seamline artifacts.

\paragraph{Quantitative Results.}
The quantitative comparison is shown in Tab.~\ref{tab:quantitative-comparisons}.
It can be seen that the proposed method achieves the highest LPIPS score in both REDS30 and SynVideo30, indicating the high fidelity of our method for synthetic videos. However, our method show inferior performance in PSNR and SSIM. This can be explained by the fact that our method tends to generate rich and vivid details, which are not favorable to metrics that capture pixel-wise deviations (i.e. PSRN and SSIM). Instead, as a perceptual-level fidelity measure, LPIPS might be more informative to reflect the content fidelity for generative tasks.
We evaluate the visual quality of generated high-resolution videos through comparisons with competing methods on MUSIQ and DOVER metrics, demonstrating our approach's superiority. This performance reflects contributions from both the base model’s capabilities and the effectiveness of our method in leveraging full-frame generation for patch-level detail synthesis.
Interestingly, our approach even preserves subjective properties like aesthetic quality from the base model, as evidenced by the aesthetic scores. This highlights our framework’s ability to inherit and maintain critical perceptual characteristics during the enhancement process.

\subsection{Ablation Study}
\label{subsec:ablation}

We conduct ablation studies on key designs. Considering the limited space, we provide a quantitative evaluation here and the typical visual results are compared in the supplementary material.

\vspace{-0.5em}
\paragraph{Global Branch.}
To validate the effectiveness of the global branch, We build a baseline by removing the global branch from our method. As shown in Tab.~\ref{tab:ablation-study}, removing the global branch results in significant performance deterioration, leading to worse DOVER, MUSIQ, and \textit{aesthetic score}. 
Besides, to check the necessity of using the patch location embeddings, we build a baseline by removing the patch position information from the global branch input. As shown in Tab.~\ref{tab:ablation-study}, after incorporating the location embedding, we obtain an increase of DOVER from 0.574 to 0.590, which demonstrates the effectiveness of utilizing the location mask as the patch position information. 

\begin{table}[!t]
    \centering
    \setlength{\tabcolsep}{3pt}
    \begin{tabular}{l|ccccc}
    \toprule
    Component          & DOVER $\uparrow$ & MUSIQ $\uparrow$ & Aesthetics $\uparrow$ \\
    \midrule
    w/o global branch  & 0.502            & 46.074           & 0.589                 \\
    w/o LoRA           & 0.582            & 50.496           & 0.600                 \\
    w/o location embed & 0.574            & 50.084           & 0.601                 \\
    w/o fixed prompt   & 0.562            & 48.133           & 0.597                 \\
    Ours               & \textbf{0.590}   & \textbf{50.559}   & \textbf{0.602}        \\
    \bottomrule
    \end{tabular}\vspace{-0.5em}
    \caption{Ablation study of different model components on VideoGen30 test dataset. The \textbf{bold} numbers indicate the best.}
    \label{tab:ablation-study}\vspace{-1.5em}
\end{table}

\vspace{-0.8em}
\paragraph{Text Prompts.}
Text prompts are used in the base model and the conditional global branch. As shown in Tab.~\ref{tab:ablation-study}, we observe that replacing the text prompt with a fixed prompt leads to superior results. We conjecture that the use of text prompts in patch condition branch brings about a mismatch between the video patch and the global semantic information, which in turn destroys the dynamic visual prior in the pre-trained model aligned to the text during training.
\vspace{-0.8em}
\paragraph{Base model finetuning.}
Since the data distribution of local patches deviates from the full view of the range of common videos, we propose to fine-tune the base model instead of freezing it. It can be seen in Tab.~\ref{tab:ablation-study} that adding LoRA results in improved performance across all metrics, which demonstrates the effectiveness of utilizing LoRA to adapt to the distribution of local video patches.

\subsection{Analysis and Discussion}
\label{subsec:discussion}

\begin{figure}[!t]
    \centering
    \includegraphics[width=\linewidth]{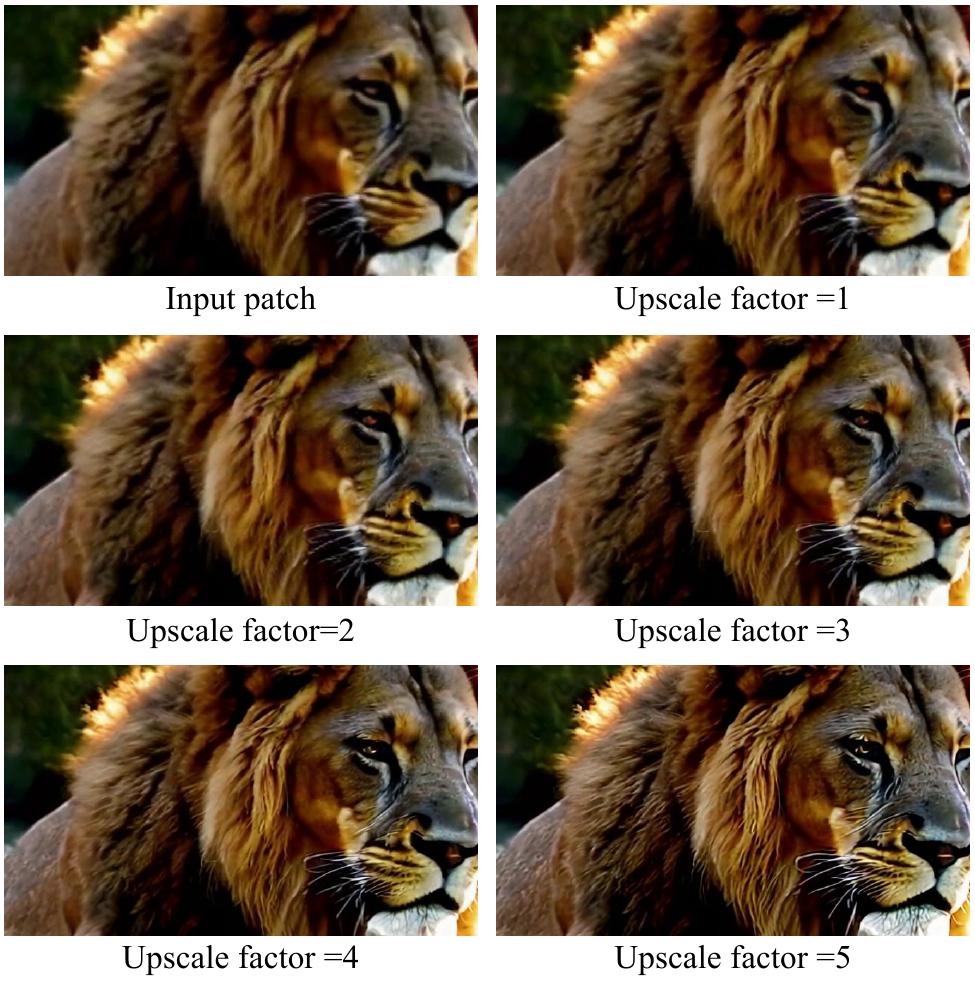}\vspace{-0.5em}
    \caption{Comparison of different up-sampling scales. Note that the upscale factor is uniformly randomly sampled within the range of 1 to 4 during the training phase.}
    \label{fig:upscale_study}\vspace{-0.5em}
\end{figure}

We would like to further study some characteristics of our proposed method, including detail generation magnitude, robustness to patch semantics, and computation efficiency.
\vspace{-0.8em}
\paragraph{Detail generation magnitude.}
The input and output sizes of our VSR model are the same and our model can synthesize different level of details. In our method, a scale factor is used to guide the model to generate reasonable details that should be presented on the input video if it is captured normally by the camera at that resolution/size. So, the scale factor actually plays the role of affecting the magnitude of details. Here we illustrate the effects of using different scale factors as conditions. In Fig.~\ref{fig:upscale_study}, increasing the scale factor progressively adds more high-resolution details to the results, which demonstrates the flexibility of our method to generate results with different detail magnitudes.  

\begin{table}[!t]
    \centering
    \begin{tabular}{lcc}
        \toprule
        Methods          & Time (s) $\downarrow$ & Memory (G) $\downarrow$ \\
        \midrule
         LaVie-SR        & 2261        & 68 \\
         Upscale-A-Video & 2743        & 47 \\
         VEnhancer       & 1562        & 62 \\
         Ours           & \textbf{680} & \textbf{40} \\
        \bottomrule
    \end{tabular}\vspace{-0.5em}
    \caption{Computation efficiency for 2K video generation.}
    \label{tab:computation-efficiency}\vspace{-0.5em}
\end{table}

\begin{figure}[!t]
    \centering
    \includegraphics[width=\linewidth]{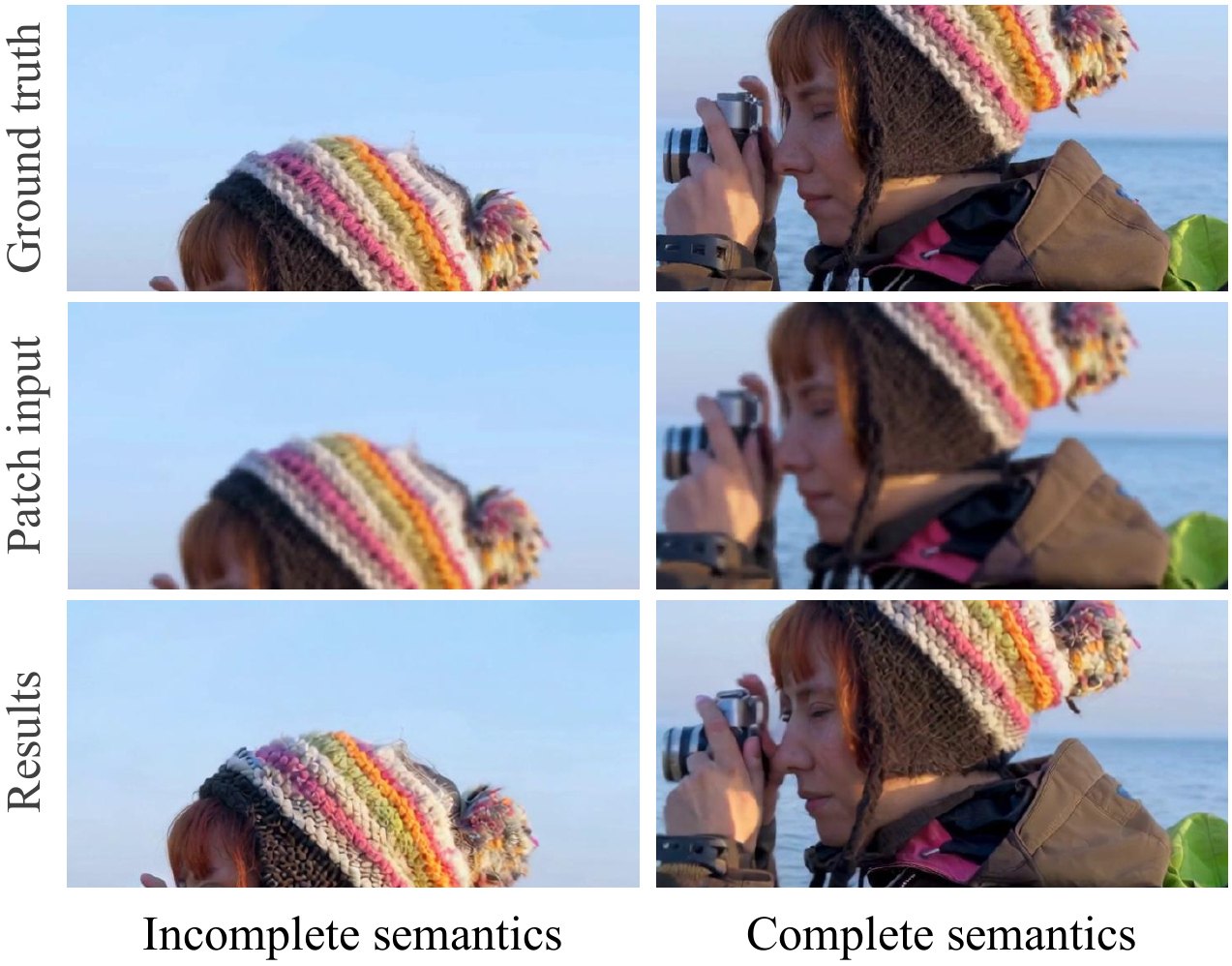}\vspace{-0.5em}
    \caption{Robustness against patch semantic incompleteness.}
    \label{fig:semantic_robustness}\vspace{-0.8em}
\end{figure}

\vspace{-0.8em}
\paragraph{Robustness to Patch Semantics.}
As stated above, our approach will partition the input video into patches in a content-agnostic manner. So, some patches can contain incomplete or even unclear semantics, which are more challenging for the model with the generative reasonable details. It is interesting to study whether the performance varies across patches with diverse semantic quality. It can be seen in Fig.~\ref{fig:semantic_robustness} that our model can successfully handle the video patches with incomplete semantics. As the global semantic tokens are extracted and input to the base model using global cross-attention modules, the model learns to utilize global semantic information to complete the semantics in the local video patch, which improves the robustness of our model against incomplete semantics.
\vspace{-0.8em}
\paragraph{Computation efficiency.}
Our method is more efficient than existing pre-trained T2V-based VSR methods because the self-attention is conducted within patches. Here presents a simple deduce of the computation complexity against patch size: given a video of $n$ tokens and each patch of $m$ tokens, the patch number is $k=\frac{n}{m}$ and the inter-token computation times are $k*(\frac{n}{k})^2=n*m$. Since the base model we adopted uses fixed $512 \times 512$ resolution, our patch size follow this size in our experiments.
The inference time and memory usage of generating a 2K resolution video is illustrated in Tab.~\ref{tab:computation-efficiency}. It can be seen that, after dividing the input video into patches, the computation burden is significantly reduced, further promoting the applications of generative VSR models. Thanks to the computation efficiency and the inference flexibility, our method even support 4K video generation, which are shown in the supplementary materials.

\paragraph{Limitation.}
Our method has limitations in two aspects. Firstly, it experiences noticeable performance degradation when applied to real-world low-resolution images, due to the lack of specialized data augmentation in our training pairs. Since the primary focus of this paper is to address the resolution limitations of using pre-trained base models for VSR, extending our approach to handle real-world noise remains an interesting direction for future work. Secondly, our approach inherits an iterative inference scheme from the base model, which introduces efficiency issues for practical applications. For VSR, the low-resolution input video provides significant opportunities for algorithm optimization to reduce the number of inference steps, which will be another area of future research.
\section{Conclusion}
\label{sec:conclusion}


In this paper, we present the first exploration of utilizing a pre-trained T2V base model for patch-level video super-resolution (VSR). To accomplish this, we propose an effective dual-branch adapter consisting of a patch condition branch and a global context branch. The global context branch is crucial for bridging the gap between full-video generative capabilities and patch-level detail synthesis. Additionally, our proposed multi-patch joint modulation scheme achieves consistent results across patches. Thanks to the flexibility of our PatchVSR approach, it outperforms existing VSR methods in generating faithful details while maintaining high computational efficiency. We hope this work will inspire further research in this area.
\newpage

\section*{Acknowledgement}
This work was supported by Kuaishou Technology and National Natural Science Foundation of China (NSFC) under Grant No.62406167, U24B6012.

\clearpage
\setcounter{page}{1}
\maketitlesupplementary

\section{Model Architecture}
\begin{figure*}[t]
    \centering
    \includegraphics[width=\linewidth]{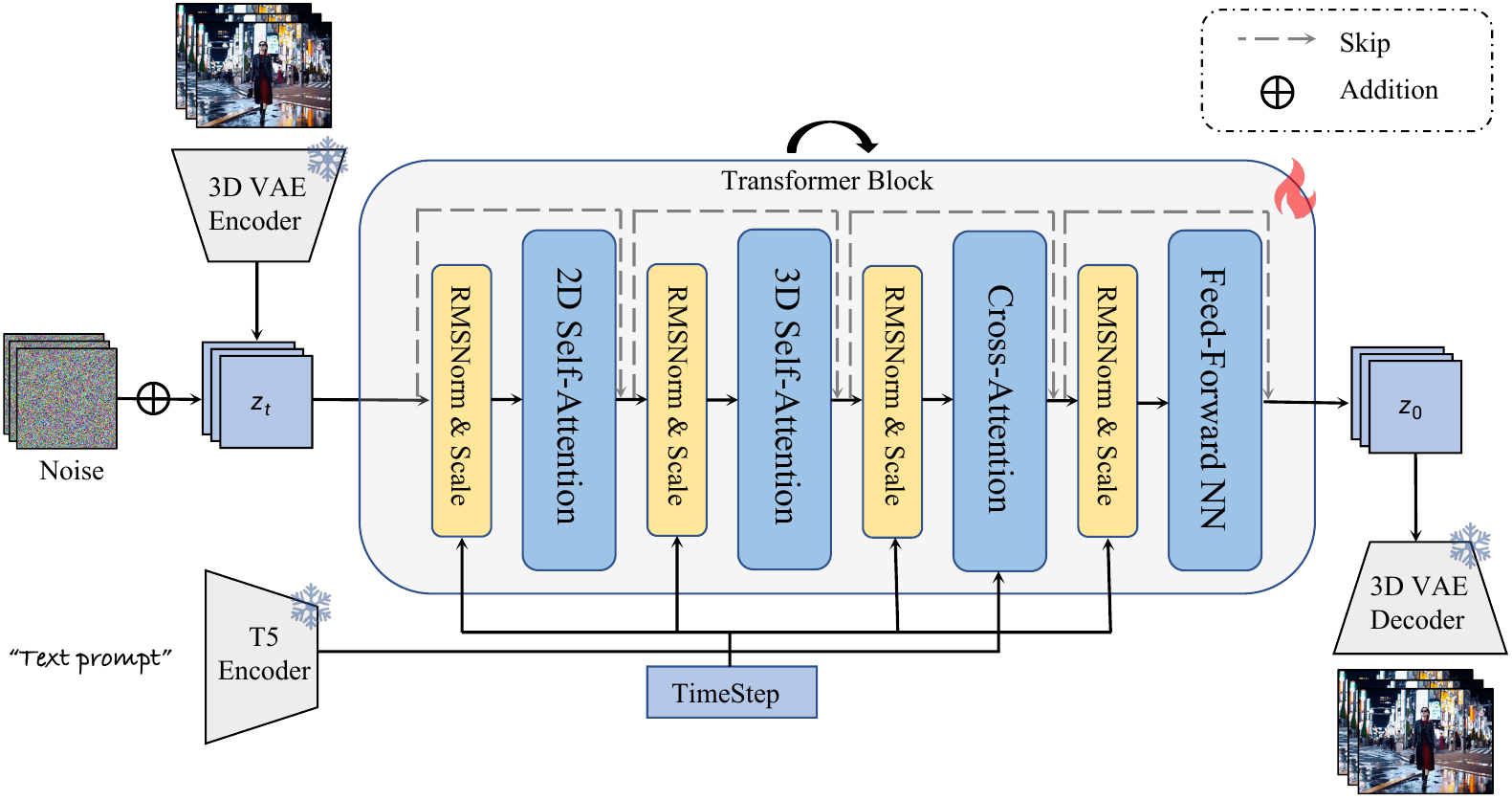}
    \caption{An overview of the architecture of our base model.}
    \label{fig:base-model}
\end{figure*}

\paragraph{Base Model.} The architecture of our base model is shown in Fig.~\ref{fig:base-model}. Our PatchVSR is built upon a pre-trained DiT-based video diffusion model, which comprises four main components: spatial self-attention (SSA), spatial cross-attention (SCA), temporal self-attention (TSA) and feed-forward network (FFN). Text prompts, time step and other micro conditions (aspect ratio, FPS, etc) are injected via the modulation mechanism \cite{peebles2023scalable}. The pre-trained model is trained on 77-frames 512$\times$512 high-quality video data with diverse aspect ratio using NaViT \cite{dehghani2024patch}. During training, we additionally fine-tune the base model using a LoRA module for all four components. We have also introduced global cross-attention (GCA) modules in SCA blocks to fuse global context feature with the backbone feature.

\paragraph{Patch Condition Branch.} The patch condition branch comprises several transformer blocks, which shares the same architecture as base model. After projected to the latent space by 3D VAE, the latent is diffused with time step $250$ (where $1000$ step denotes pure noises in our noise scheduling) before input to the adapter in each denoise time step during inference. The parameters of patch condition branch are initialized from base model and the layer indices are chosen at equal intervals following \cite{lin2024ctrl}.

\paragraph{Global Context Branch.} To provide initialization weight, we also follow the structure of the pre-trained model and sample the layer indices from it at equal intervals. To obtain high-level global semantic token, we design a transformer-based encoder architecture to compress the token number layer to one quarter layer by layer. The transformer blocks are evenly split for each token level. The token number is reduced for the output of the transformer blocks using the patchify operation \cite{peebles2023scalable}. The input to the global context branch is the latent of the resized full video encoded by the 3D VAE of the base model. We also diffuse the latent with time step $150$ for noise augmentation. The choice of time step comes from a fact that the noise added to the full video will have a greater impact on local areas, and we set a smaller noise to maintain the noise intensity received by the corresponding local areas in patch condition branch and global context branch. Before input to the transformer blocks, the binary map is concatenated with the latent at the channel dimension, which indicates the target patch location within it.

\section{Additional Results}

\subsection{Visual Comparisons on 2K Full Videos}

\begin{figure*}
    \centering
    \includegraphics[width=\linewidth]{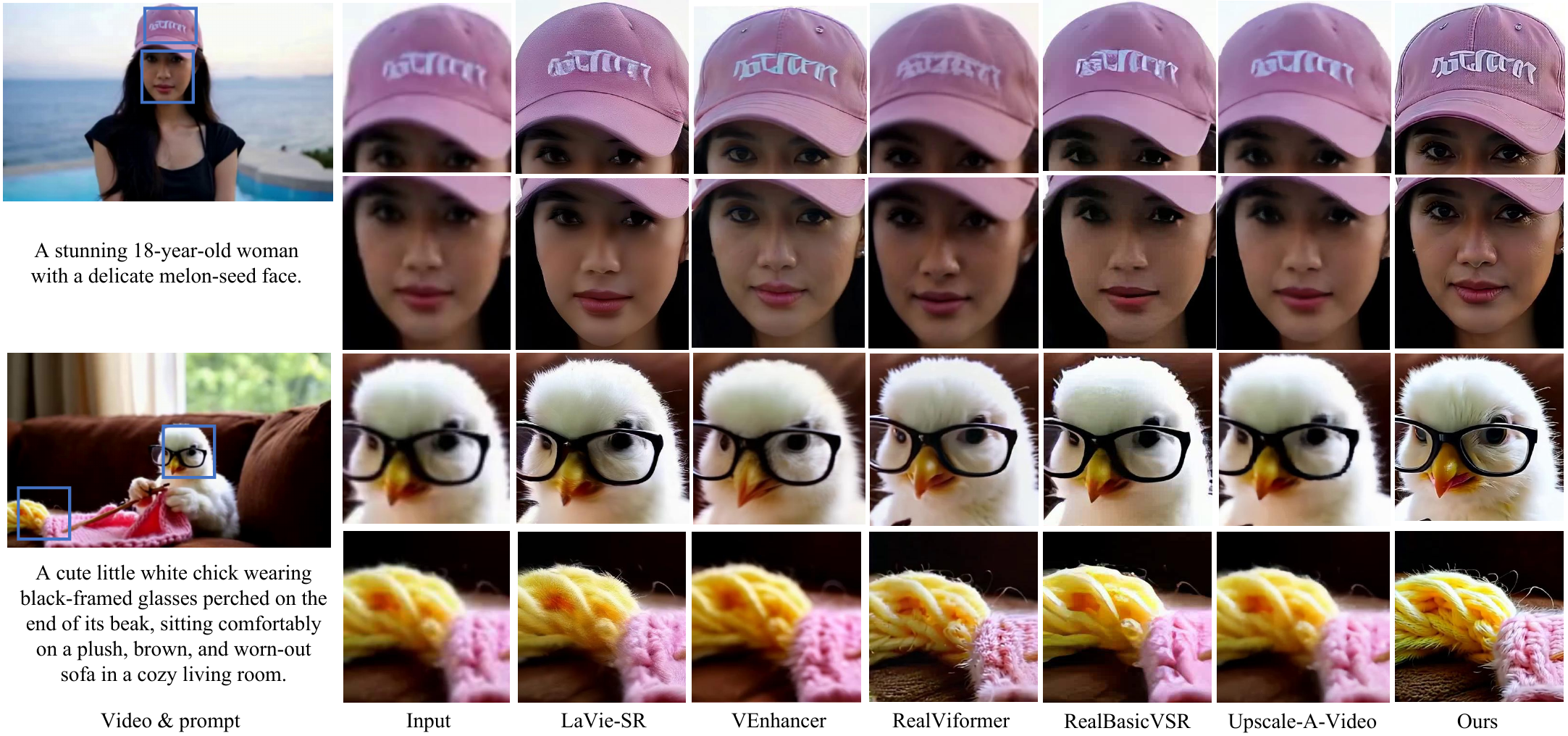}
    \caption{Comparisons of 2K full videos on VideoGen30 (test). \textbf{Zoom in for best view.}}
    \label{fig:supp-aigc-comparison}
\end{figure*}

\paragraph{AIGC videos.} The comparisons of 2K AIGC videos are shown in Fig.~\ref{fig:supp-aigc-comparison}. The results show that our method can effectively generalize to AI-generated videos, generating rich local details and textures while guaranteeing high fidelity.

\begin{figure*}
    \centering
    \includegraphics[width=\linewidth]{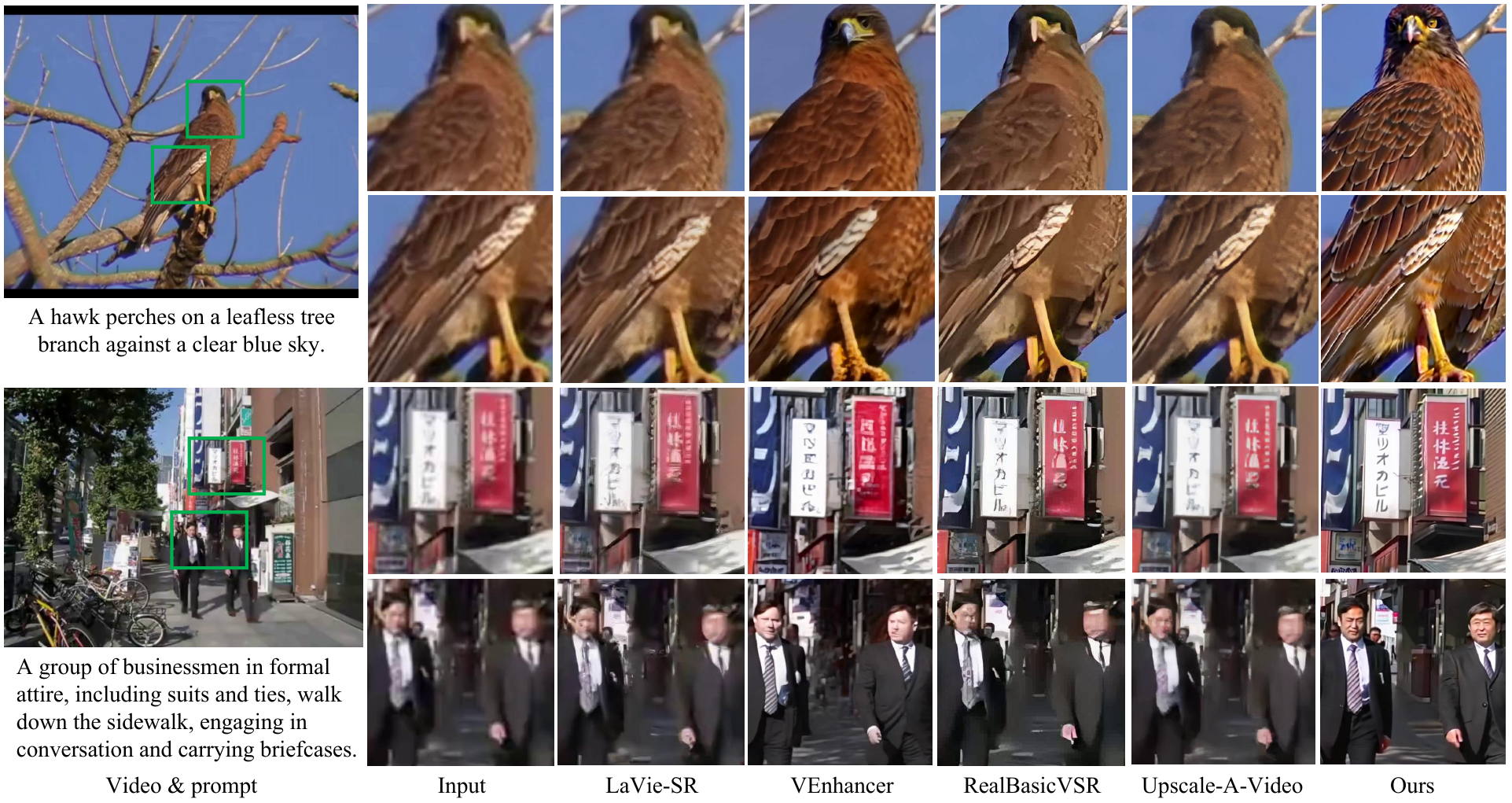}
    \caption{Qualitative comparisons on VideoLQ test set.}
    \label{fig:qualitative-videolq}
\end{figure*}

\paragraph{Real-world Videos} To adapt our model for real-world degraded videos, we further finetuned our model on a subset of our training data that involves degradation simulation pipeline like \cite{chan2022investigating}. The qualitative comparison and quantitative evaluation are shown in Fig.~\ref{fig:qualitative-videolq} and Tab.~\ref{tab:quantitative-videolq} respectively.

\begin{table*}
    \centering
    \resizebox{0.8\linewidth}{!}{
    \begin{tabular}{c|cccc|c}
       \toprule
		Metrics & RealBasicVSR & LaVie-SR & Upscale-A-Video & VEnhancer & Ours \\
		\midrule
        DOVER $\uparrow$ & \underline{0.497} & 0.371 & 0.350 & 0.411 & \textbf{0.501} \\
        MUSIQ $\uparrow$ & \textbf{52.074} & 36.204 & 24.599 & 35.157 & \underline{37.342} \\
        Aesthetics $\uparrow$ & \underline{0.531} & 0.521 & 0.526 & 0.522 & \textbf{0.549} \\
        \bottomrule
    \end{tabular}
    }
    \caption{Quantitative results on VideoLQ test set.}
    \label{tab:quantitative-videolq}
\end{table*}

\subsection{Quantitative Comparisons on VideoScore}
As evaluation by additional benchmark VideoScore \cite{he2024videoscore}, Tab.~\ref{tab:videoscore-videogen30} tells the best performance of PatchVSR on \textit{temporal consistency} and \textit{dynamic degree} among all compared methods.

\begin{table*}
    \centering
    \resizebox{0.8\linewidth}{!}{
    \begin{tabular}{c|cccc|c}
       \toprule
		Metrics & RealBasicVSR & LaVie-SR & Upscale-A-Video & VEnhancer & Ours \\
		\midrule
        Visual Quality $\uparrow$ & 2.669 & \underline{2.749} & 2.685 & 2.686 & \textbf{2.811} \\
        Temporal Consistency $\uparrow$ & 2.669 & \underline{2.848} & 2.818 & 2.844 & \textbf{2.897} \\
        Dynamic Degree $\uparrow$ & 2.481 & \underline{2.488} & 2.392 & 2.380 & \textbf{2.619} \\
        Text-to-video Alignment $\uparrow$ & 2.506 & \textbf{2.598} & 2.541 & \underline{2.563} & 2.490 \\
        Factual Consistency $\uparrow$ & 2.613 & \underline{2.703} & 2.673 & 2.686 & \textbf{2.768} \\
        \bottomrule
    \end{tabular}
    }
    \caption{VideoScore evaluation on VideoGen30 test set.}
    \label{tab:videoscore-videogen30}
\end{table*}

\subsection{Visualization on 4K Full Videos}
\begin{figure*}
    \centering
    \includegraphics[width=\linewidth]{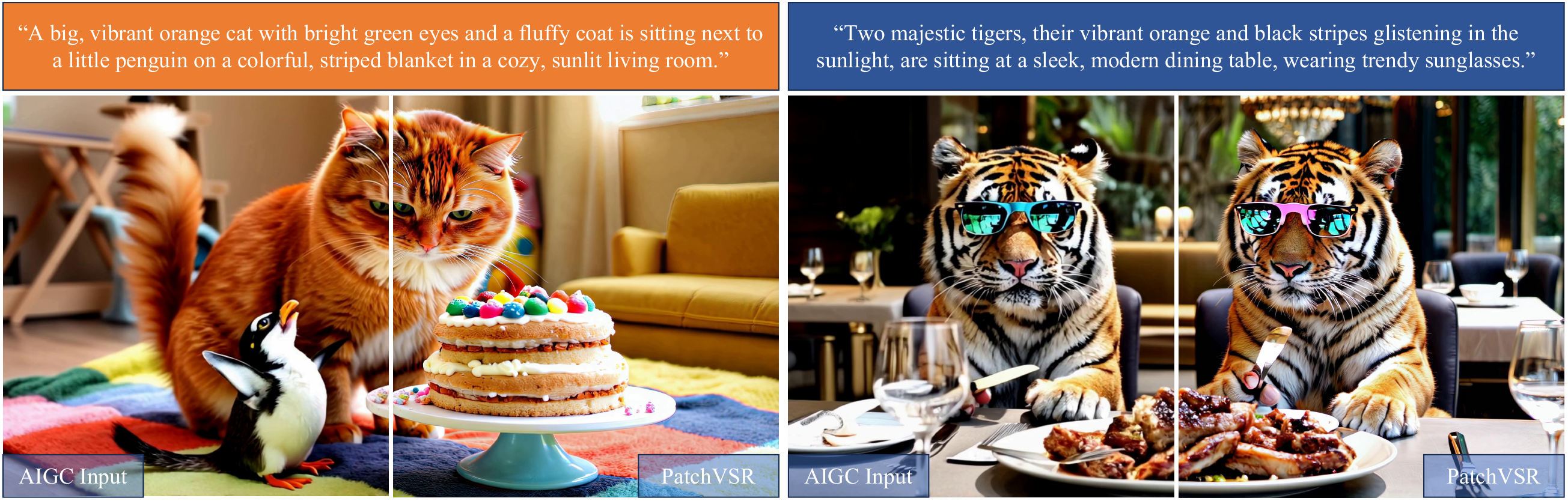}
    \caption{Visualization of 4K full videos based on 720P input. \textbf{Zoom in for best view.}}
    \label{fig:4k-aigc}
\end{figure*}

We have also performed 4K video super-resolution using 720P input video. Results can be seen in Fig.~\ref{fig:4k-aigc}. It can be seen that our method successfully generate 4K-resolution video with high clarity and fine details.

\subsection{Ablation Study}

\begin{figure*}[h]
    \centering
    \includegraphics[width=\linewidth]{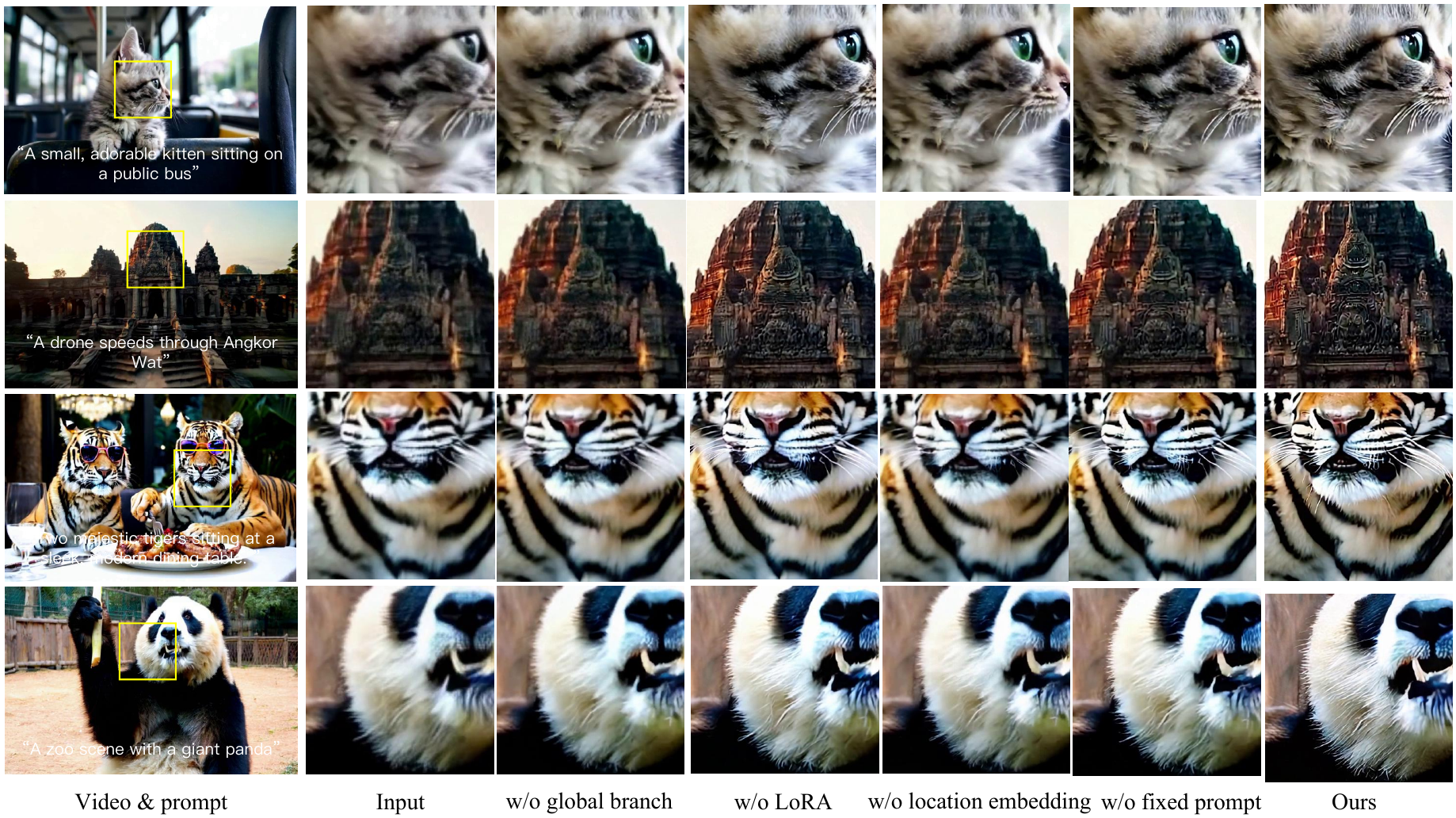}
    \caption{The qualitative comparisons of the ablation studies.}
    \label{fig:supp-ablation-study}
\end{figure*}

\paragraph{Qualitative Comparisons.} The visual results of the ablation studies are shown in Fig.~\ref{fig:supp-ablation-study}. While the use of local condition branch provides basic low-frequency information such as region structure and color, it lacks the ability to generate accurate high-frequency details. Besides, the lack of guidance from location embedding can make the global semantic tokens provided by global context branch less accurate, preventing the model from generating further high-definition details and textures. We also observe that removing LoRA makes the local high-frequency details of the generated results missing and inaccurate, consistent with our view that adding LoRA can facilitate the matching of pre-trained model distribution with patch video distribution.

\paragraph{Fidelity Evaluation.} We provide the comparison in Tab.~\ref{tab:fidelity-synvideo30}. It shows that, although introducing global branch brings slight fidelity degradation, using location embedding will partly mitigate it. Also, using global prompt only will lead to the mismatch between global context and video patch, which severely affect the performance.

\begin{table*}
    \centering
    \resizebox{0.8\linewidth}{!}{
    \begin{tabular}{c|cccc|c}
       \toprule
		Metrics & w/o global branch & w/o LoRA & w/o location embedding & w/o fixed prompt & Ours \\
		\midrule
        PSNR $\uparrow$ & \textbf{31.245} & 30.744 & 30.563 & 30.142 & \underline{30.857} \\
        SSIM $\uparrow$ & \textbf{0.744} & \underline{0.736} & 0.707 & 0.688 & 0.732 \\
        LPIPS $\downarrow$ & \textbf{0.176} & 0.185 & 0.195 & 0.201 & \underline{0.183} \\
        \bottomrule
    \end{tabular}
    }
    \caption{Fidelity evaluation on SynVideo30 test set.}
    \label{tab:fidelity-synvideo30}
\end{table*}

\subsection{Stitched Schemes}
\begin{figure*}
    \centering
    \includegraphics[width=\linewidth]{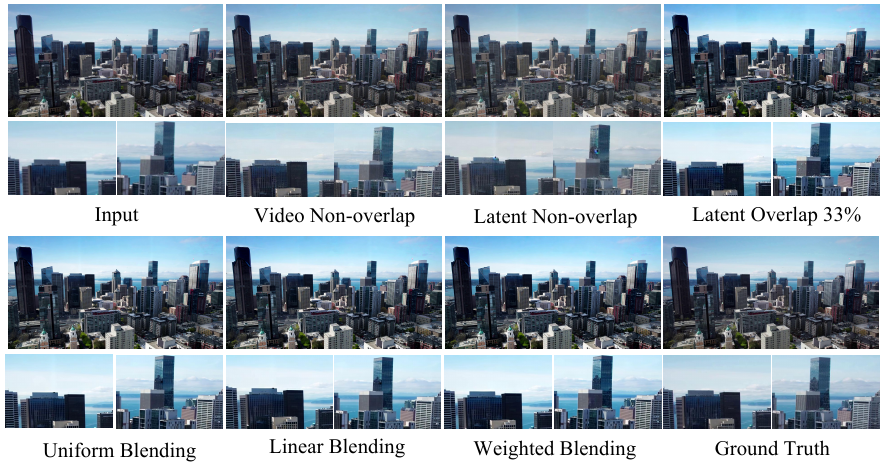}
    \caption{Comparisons of stitch schemes of 2K full video.}
    \label{fig:stitch-comparison}
\end{figure*}

Although original multi-patch modulation successfully solves the seam problem of 2K full videos, it inevitably occurs black holes due to feature mismatch at the intersection of multiple mismatches. To tackle this problem, we further propose a weighted multi-patch modulation technique to obtain smooth feature. Specifically, for the four pixels adjacent to the black hole, their features in latent space all originate from three patches (including one primary patch and two auxiliary patches). When a uniform mixing strategy is applied to the features of the auxiliary patches, any two of these four points have two patches of different origins, leading to a severe black hole phenomenon. After applying the linear blending strategy, only one patch source is different while the other different patch source has a weight of 0 here, thus attenuating the black hole phenomenon in the stitched video. However, due to the feature mismatch of the primary patch, the black hole still occurs. To eliminate the black hole, we manually set larger weight for the feature of the shared patch (one of the auxiliary patches) to amplify its feature, which results in smooth feature transition, thus preventing from generating black holes. The comparisons of different stitch schemes are shown in Fig.~\ref{fig:stitch-comparison}. The results show that when the blending weight is set to 5.0, the seam position transitions smoothly and the black holes are effectively removed for all examples.

\end{document}